\pdfoutput=1
\documentclass{llncs}
\usepackage{makeidx}  
\usepackage{color,colortbl}
\usepackage[pdftex]{graphicx}
\usepackage{bm}
\usepackage{caption}
\usepackage{subcaption}
\usepackage{multirow}
\usepackage{float}
\usepackage{eqparbox}
\usepackage[cmex10]{amsmath}
\usepackage{tikz,pgfplots,pgfplotstable}
\usetikzlibrary{arrows,shapes,positioning,shadows,trees}
\usetikzlibrary{pgfplots.groupplots}
\usetikzlibrary{backgrounds, calc, hobby}
\usetikzlibrary{shapes.geometric}
\definecolor{mygray}{gray}{0.6}
\newcommand{\numcodes}{11,881}
\begin{document}
\frontmatter          
\pagestyle{headings}  
%

%
%
\title{Hospital Readmission Prediction - Applying Hierarchical Sparsity Norms for Interpretable Models}
\titlerunning{Exploiting Hierarchy in Disease Codes}  
%
\author{Jialiang Jiang%
\thanks{\{jjiang6,hewner,chandola\}@buffalo.edu, State University of New York at Buffalo}\inst{1}%
\and Sharon Hewner\footnotemark[1]\inst{2}
\and Varun Chandola\footnotemark[1]\inst{1}
}
\authorrunning{Jialiang Jiang et al.}
%
\tocauthor{Jialiang Jiang, Sharon Hewner, Varun Chandola}
\institute{Department of Computer Science and Engineering, State University of New York at Buffalo \\
\and School of Nursing, State University of New York at Buffalo}

\maketitle              

\begin{abstract}
Hospital readmissions have become one of the key measures of healthcare quality. Preventable readmissions have been identified as one of the primary targets for reducing costs and improving healthcare delivery. However, most data driven studies for understanding readmissions have produced black box classification and predictive models with moderate performance, which precludes them from being used effectively within the decision support systems in the hospitals. 
In this paper we present an application of structured sparsity-inducing norms for predicting readmission risk for patients based on their disease history and demographics. Most existing studies have focused on hospital utilization, test results, etc., to assign a readmission label to each episode of hospitalization. However, we focus on assigning a readmission risk label to a patient based on their disease history. Our emphasis is on interpreting the models to improve the understanding of the readmission problem. To achieve this, we exploit the domain induced hierarchical structure available for the disease codes which are the features for the classification algorithm. We use a tree based sparsity-inducing regularization strategy that explicitly uses the domain hierarchy. The resulting model not only outperforms standard regularization procedures but is also highly sparse and interpretable. We analyze the model and identify several significant factors that have an effect on readmission risk. Some of these factors conform to existing beliefs, e.g., impact of surgical complications and infections during hospital stay. Other factors, such as the impact of mental disorder and substance abuse on readmission, provide empirical evidence for several pre-existing but unverified hypotheses. The analysis also reveals previously undiscovered connections such as the influence of socioeconomic factors like lack of housing and malnutrition. 
The findings of this study will be instrumental in designing the next generation decision support systems for preventing readmissions.

\end{abstract}

\section{Introduction}
\label{sec:introduction}
Hospital readmissions are prevalent in the healthcare system and contribute significantly to avoidable costs. In United States, recent studies have shown that the 30-day readmission rate among the Medicare beneficiaries\footnote{A federally funded insurace program representing 47.2 \% (\$182.7 billion) of total aggregate inpatient hospital costs in the United States\protect\cite{Pfunter:2013}.} is over 17\%, with close to 75\% of these being avoidable~\cite{Mpac:2007}, with an estimated cost of \$15 Billion in Medicare spending. Similar alarming statistics are reported for other private and public insurance systems in the US and other parts of the world. In fact, management of care transitions to avoid readmissions has become a priority for many acute care facilities as readmission rates are increasingly being used as a measure of quality~\cite{Conway:2011}.

Given that the rate of avoidable readmission has now become a key measure of the quality of care provided in a hospital, there have been increasingly large number of studies that use healthcare data for understanding readmissions. Most existing studies have focused on building models for predicting readmissions using a variety of available data, including patient demographic and social characteristics, hospital utilization, medications, procedures, existing conditions, and lab tests~\cite{Futoma:2015,Choudhry:2013,Donze:2013}. Other methods use less detailed information such as insurance claim records~\cite{Yu:2013,He:2014}. Many of these methods use machine learning methods, primarily Logistic Regression, to build classifiers and report consistent performance. In fact, most papers about readmission prediction report AUC scores in the range of 0.65-0.75.

While the predictive models show promise, their moderate performance means that they are still not at a stage where hospitals could use them as ``black-box'' decision support tools. However, given that the focus of these predictive models has been on performance, the models themselves are not easily interpretable to provide actionable insights to the decision makers. In this paper, we present a methodology to infer such insights from healthcare data in the context of readmissions. We build a logistic regression based classifier to predict if a patient is likely to be readmitted based on their disease history available from insurance records. We use sparsity inducing regularizers in our predictive model to promote interpretability. In particular, we show that by exploiting the hierarchical relationship between disease codes using the {\em tree-structured hierarchical group regularization}~\cite{Zhao:2009}, we are able to learn a predictive model that outperforms all other types of sparsity inducing norms. Moreover, the tree-structured norm allows us to incorporate the rich semantic information present in the disease code taxonomy into the model learning, yielding highly interpretable models.

By analyzing the model trained on claims data obtained from the New York State Medicaid Warehouse (MDW), we infer several important insights to improve the understanding of readmissions. Some of our findings conform to existing beliefs, for example, the importance of bacterial infections during hospital stay. Other findings provide empirical evidence to support existing hypotheses amongst healthcare practitioners, for example, the effect of the type of insurance on readmissions~\cite{Hewner:2014}. Most interesting findings from our study reveal surprising connections between a patient's non-disease background and the risk of readmission. These include behavioral patterns (mental disorders, substance abuse) and socio-economic background.
\tikzset{
  fontscale/.style = {font=\footnotesize},
  basic/.style  = {align=center, draw, text width=2cm, drop shadow, font=\sffamily, rectangle, fontscale},
  root/.style   = {basic, rounded corners=2pt, thin, align=center,
  fill=green!30,text width=10em},
  level 2/.style = {basic, rounded corners=6pt, thin,align=center, fill=green!60,
  text width=8em},
  level 3/.style = {basic, thin, align=left, fill=pink!60, text width=7.5em}
}

\begin{figure*}[tbp]
  \centering
\begin{tikzpicture}[
    level 1/.style={sibling distance=32mm},
    edge from parent/.style={->,draw},
  ]

  \node[root] {\bf 2015 ICD-9-CM}
  child {node[level 2] (c1) {Infectious \& Parasitic}}
  child {node[level 2] (c2) {Neoplasms}}
  child {node (c0) {$\ldots$}}
  child {node[level 2] (c3) {Injury \& Poisoning}};

  \begin{scope}[every node/.style={level 3}]
    \node [below of = c1, xshift=8pt] (c11) {Intestinal Infections};
    \node [below of = c11] (c12) {Tuberculosis};
    \node [below of = c12] (c13) {Zoonotic Bacterial Infections};
    \node [below of = c13] (c14) {$\ldots$};

    \node [below of = c2, xshift=8pt] (c21) {Malignant (Lip, Oral Cavity, $\ldots$)};
    \node [below of = c21] (c22) {Malignant (Digestive)};
    \node [below of = c22] (c23) {$\ldots$};
    \node [below of = c3, xshift=8pt] (c31) {External cause status};
    \node [below of = c31] (c32) {Activity};
    \node [below of = c32] (c33) {Railway Accidents};
    \node [below of = c33] (c34) {$\ldots$};

  \end{scope}

  \draw[->] (c1.west) |- (c11.west);
  \draw[->] (c1.west) |- (c12.west);
  \draw[->] (c1.west) |- (c13.west);
  \draw[->] (c1.west) |- (c14.west);

  \draw[->] (c2.west) |- (c21.west);
  \draw[->] (c2.west) |- (c22.west);
  \draw[->] (c2.west) |- (c23.west);

  \draw[->] (c3.west) |- (c31.west);
  \draw[->] (c3.west) |- (c32.west);
  \draw[->] (c3.west) |- (c33.west);
  \draw[->] (c3.west) |- (c34.west);
\end{tikzpicture}
\caption{A sample of ICD9-CM classification.}
\label{fig:icd9}
\end{figure*}
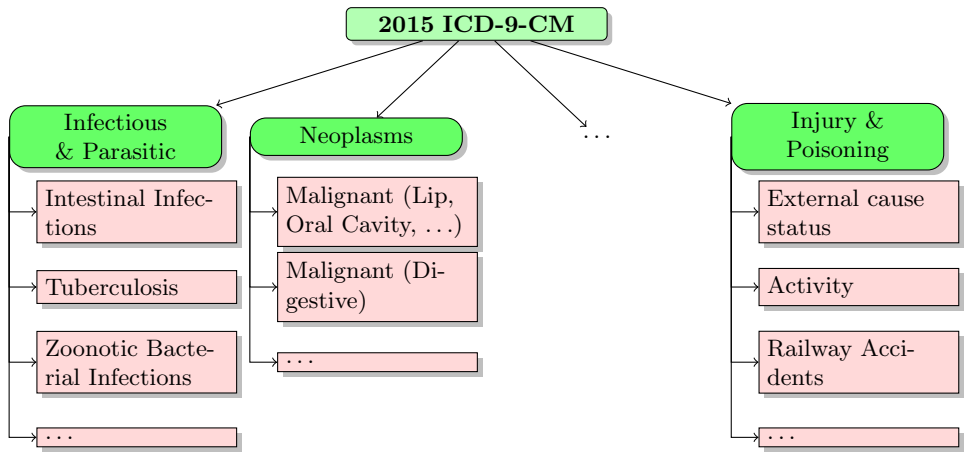

We believe that such findings can have a significant impact on how healthcare providers develop effective strategies to reduce readmissions. At present, the healthcare efforts in this context have been two fold. First is the effort to improve the quality of care within the hospital and the second is to develop effective post-discharge strategies such as telephone outreach, community-based interventions, etc. The results from this study inform the domain experts on both fronts. 
\paragraph{\bf Organization}
The rest of the paper is organized as follows. We review existing literature on readmission prediction in Section~\ref{sec:related}. We describe the data used for our experiments in Section~\ref{sec:data} and formulate the machine learning problem in Section~\ref{sec:problem}. We discuss the classification methodology in Section~\ref{sec:methodology}. The results are presented in Section~\ref{sec:results}. The analysis of the resulting model in the context of readmissions is provided in Section~\ref{sec:discussions}. We present analysis of experiments done on subpopulations corresponding to major chronic diseases in Section~\ref{sec:subpop}.

\section{Related Work}
\label{sec:related}
Coincident with the rising importance of readmissions in reducing healthcare costs, there have been several papers that use clinical and insurance claims information to build predictive models for readmissions. We refer the readers to a recent survey on the topic~\cite{Kansagara:2011} for a comprehensive review. Most of these models use machine learning models such as Logistic Regression~\cite{Futoma:2015,Choudhry:2013,Donze:2013,Niu:2013} and Support Vector Machines~\cite{Yu:2013}. Futoma et. al~\cite{Futoma:2015} provide a comparison of several machine learning methods including Logistic Regression, SVM, Random Forests for readmission prediction using features such as the diagnosis codes, procedure codes, demographics, patient's hospitalization history. However, the focus of most of these papers has been on improving accuracy of the classifier and not on interpreting the models to improve the understanding of the readmission problem. Moreover, many of these studies have either focused on a specific patient cohort or patient data from one or few hospitals~\cite{Yu:2013}. For example, there have been several studies that focus on patients with acute heart conditions~\cite{Amarasingham:2010}. Recently, healthcare community has begun to study the impact of behavioral and socioeconomic factors on readmissions~\cite{Hewner:2015}. However, none of the data driven predictive models exploit this aspect, primarily because such data is challenging to obtain. However, in this paper we show that the diagnosis codes in the claims data contains valuable non-disease information about the patient which can be leveraged to better understand the readmission issue. 
Finally, the hierarchical relationship has never been exploited for building predictive models for readmission. Singh, et. al,~\cite{Singh:2014} have presented a similar approach in the context of predicting disease progression, however, the authors focus on using the disease hierarchy to come up new features that are fed into the classifier.

\section{Data}
\label{sec:data}
The data is obtained from the New York State Medicaid Warehouse (MDW). Medicaid is a social health care program for families and individuals with low income and limited resources. We analyzed four years (2009--2012) of claims data from MDW. The claims correspond to multiple types of health utilizations including hospitalizations, outpatient visits, etc. While the raw data consisted of 4,073,189 claims for 352,716 patients, we only included the patients in the age range 18--65 with no obstetrics related hospitalizations. The number of patients with at least one hospitalization who satisfied these conditions were 11,774 and had 34,949 claims.

For each patient we have two types of information. First type of information includes demographic attributes (age, gender) and the type of insurance. The second type of information is a patient history extracted from four years of claims data represented as a binary vector that indicates if the patient was diagnosed with a certain disease in the last four years.
\paragraph{\bf Insurance Plan Information:}
Patients covered through Medicaid insurance can enroll into one of the two plan options. First option is to enroll in a {\em Managed Care Organization} (MCO). The MCO takes care of the delivery of the medical care to the patient and gets paid per person. The other option is called {\em Fee-for-service} (FFS) in which the healthcare provider gets paid for each service performed for the patient. While there has been a gradual transition from FFS to MCO style of insurance, the quality of care and costs under each plan has always been an important issue. In the context of readmissions it is important to understand how the readmission rate is impacted by the type of plan.  
\paragraph{\bf Diagnosis Codes:}
Disease information is encoded in insurance claims using {\em diagnosis codes}. The {\em International Classification of Diseases} (ICD) is an international standard for classification of disease codes. The data used in this paper followed the ICD-9-CM classification which is a US adaptation of the ICD-9 classification. Conceptually, the ICD-9-CM codes are structured as a tree (See Figure~\ref{fig:icd9} for a sample\footnote{See \url{http://www.icd9data.com/2015/Volume1/default.htm} for complete hierarchy.}) with 19 broad disease categories at level 1. The entire tree has 5 levels and has total of 14,567 diagnosis codes. While the primary purpose of ICD taxonomy has been to support the insurance billing process, it contains a wealth of domain knowledge about the difference diseases.
\paragraph{\bf Readmission Risk Flag:}
For each patient in the above described cohort, we assign a binary flag for readmission risk. The readmission risk flag is set to 1 if the patient had {\em at least} one pair of consecutive hospitalizations within 30 days of each other in a single calendar year, otherwise it is set to 0. 

\section{Problem Statement}
\label{sec:problem}
Given a patient's demographic information and disease history, we are interested in predicting the {\em readmission risk} (binary flag) for the patient. The problem formulation is different from many existing studies~\cite{Futoma:2015}, where the focus is on assigning a readmission risk to a single hospitalization event. Our focus is on understanding the impact of socio-economic and behavioral factors on a readmission.

We denote each patient $i$ as a vector ${\bf x}_i$ consisting of 11,884 elements corresponding to \numcodes\ disease codes and three elements for age, gender, and plan type. Note that while ICD-9-CM classification contains 14,567 codes, only \numcodes\ codes are observed in the data set used in this paper.  All elements in the vector are binary except for age which takes 10 possible values corresponding to 10 equal width partitions between 18 and 65. The readmission risk flag is denoted using $y_i \in \{0,1\}$ where 1 indicates readmission risk.

From machine learning perspective our task is to learn a classifier from a training data set $\langle {\bf x}_i, y_i \rangle_{i=1}^N$ which can be used to assign the readmission risk flag to a new patient represented as ${\bf x}_*$. Note that the input vector ${\bf x}_i$ is highly sparse for this data with nearly 36 non-zeros out of total 11,884 elements on an average. 

\section{Methodology}
\label{sec:methodology}
We use a {\em logistic regression} (LR) model~\cite{Cox:1958} as the classifier, which, is the most widely used model in the context of readmission prediction~\cite{Futoma:2015}. The LR model, for binary classification tasks, computes the probability of the target $y$ to be 1 (readmission risk), given the input variables, ${\bf x}$ as:
\begin{equation}
  \label{eqn:lr}
  p(y = 1|{\bf x}) = \frac{1}{1 + \exp(-{\bm \beta}^\top{\bf x})}
\end{equation}
Where ${\bm \beta}$ is the LR model parameter (regression coefficients). We assume that ${\bf x}$ includes a constant term corresponding to the intercept. Thus ${\bm \beta}$ has the dimensionality $D+1$ where $D = 11,884$. 

The model parameter ${\bm \beta}$ are learnt from a training data set ($\langle {\bf x}_i, y_i \rangle_{i=1}^N$) by optimizing the following objective function:
\begin{equation}
  \label{eqn:lrlearn}
  {\hat{\bm \beta}} = \arg\min_{\bm \beta} \sum_{i=1}^N\log(1 + exp(-y_i{\bm \beta}^\top{\bf x}_i)) + \lambda\Omega({\bm \beta})
\end{equation}
where the first term refers to the training loss and the second terms is a regularization penalty imposed on the solution; $\lambda$ being the regularization parameter. Different forms of regularization penalties have been used in the past, including the widely used $l_1$ and $l_2$ norms~\cite{Tibshirani:1994}. While $l_2$ norm ($\Omega({\bm \beta}) = \Vert{\bm \beta}\Vert_2 = (\sum_j\beta_j^2)^{1/2}$) is typically used to ensure stable results, $l_1$ norm ($\Omega({\bm \beta}) = \Vert{\bm \beta}\Vert_1 = \sum_j\vert\beta_j\vert$) is used to promote sparsity in the solution, i.e., most coefficients in ${\bm \beta}$ are 0.

However, $l_1$ regularizer does not explicitly promote structural sparsity. Given that the features used in predicting readmission risk have a well-defined structure imposed by the ICD-9 standards, we explore regularizers that leverage this structure for model learning:
\subsubsection*{Sparse Group Regularizer}
This regularizer (also refered to as Sparse Group LASSO or SGL) assumes that the input features can be arranged into $K$ groups (non-overlapping or overlapping)~\cite{Bach:2008}. The SGL regularizer is given by:
\begin{equation}
  \label{eqn:sgl}
  \Omega({\bm \beta}) = \alpha \Vert{\bm \beta}\Vert_1 + (1- \alpha)\sum_{k=1}^{K}\Vert{\bm \beta}_{G_k}\Vert_2
\end{equation}
where ${\bm \beta}_{G_k}$ are the coefficients corresponding to the group $G_k$. The above penalty function favors solutions which select only a few groups of features (group sparsity). For the task of readmission prediction, we divide the features corresponding to \numcodes\ diagnosis codes into $19$ non-overlapping groups, based on the top level groupings in the ICD-9-CM classification (See Table~\ref{tab:icd9toplevel}). The demographic and insurance plan features are grouped into an additional group.    

\begin{table}[h]
  \centering
  {\footnotesize
    \begin{tabular}{|cp{4.8in}|}
    \hline
    1 & Infectious And Parasitic Diseases\\
    2 & Neoplasms\\
    3 & Endocrine, Nutritional And Metabolic Diseases, And Immunity Disorders\\
    4 & Diseases Of The Blood And Blood-Forming Organs\\
    5 & Mental Disorders\\
    6 & Diseases Of The Nervous System And Sense Organs\\
    7 & Diseases Of The Circulatory System\\
    8 & Diseases Of The Respiratory System\\
    9 & Diseases Of The Digestive System\\
    10& Diseases Of The Genitourinary System\\
    11& Complications Of Pregnancy, Childbirth, And The Puerperium\\
    12& Diseases Of The Skin And Subcutaneous Tissue\\
    13& Diseases Of The Musculoskeletal System And Connective Tissue\\
    14& Congenital Anomalies\\
    15& Certain Conditions Originating In The Perinatal Period\\
    16& Symptoms, Signs, And Ill-Defined Conditions\\
    17& Injury And Poisoning\\
    18& Supplementary Classification Of Factors Influencing Health Status And Contact With Health Services\\
    19& Supplementary Classification Of External Causes Of Injury And Poisoning\\
    \hline
  \end{tabular}
}
  \caption{Top level disease groups in the ICD-9-CM classification}
  \label{tab:icd9toplevel}
\end{table}
\subsubsection*{Tree Structured Group Regularizer}
This regularizer, also referred to as Tree Structured Group LASSO (TSGL), explicitly uses the hierarchical structure imposed on the features. The TSGL regularizer is given by:
\begin{equation}
  \label{eq:tsgl}
  \Omega({\bm \beta}) =\sum_{i=0}^D \sum_{j=1}^{N_i}\Vert\beta_{G_j^i}\Vert_1
\end{equation}
where $G$ denotes the tree constructed using the hierarchy of the diagnosis codes. $G^i_j$ denotes the $j^{th}$ node in the tree at the $i^{th}$ level. Thus $G^0_1$ denotes the root level, and so on. For readmission risk prediction, we consider a three internal level hierarchy with 1193 nodes at level 3, 183 nodes at level 2, and 20 nodes at level 1.

\section{Results}
\label{sec:results}
In this section we present our findings by applying logistic regression classifier for the task of readmission prediction on the MDW data described earlier. The full data set consists of 11,774 patients with 4,580 patients with readmission risk flag as {\em true} and 7,194 patients with readmission risk flag as {\em false}. We first compare the performance of the different regularization strategies to the classification task using the {\em F-measure} (harmonic mean of precision and recall for the {\em readmission = yes} class) as our evaluation metric. We also report the area under the ROC-curve (AUC) for each classifier. For each strategy, we run 10 experiments with random 60-40 splits for training and test data, respectively. The optimal values for the regularization parameter for each strategy are chosen using cross-validation. We use the MATLAB package, SLEP~\cite{Liu:2009}, for the structured regularization experiments. 
\subsection{Comparing Different Regularizations}
\begin{figure}[ht]
	\centering
	\includegraphics[width=90mm]{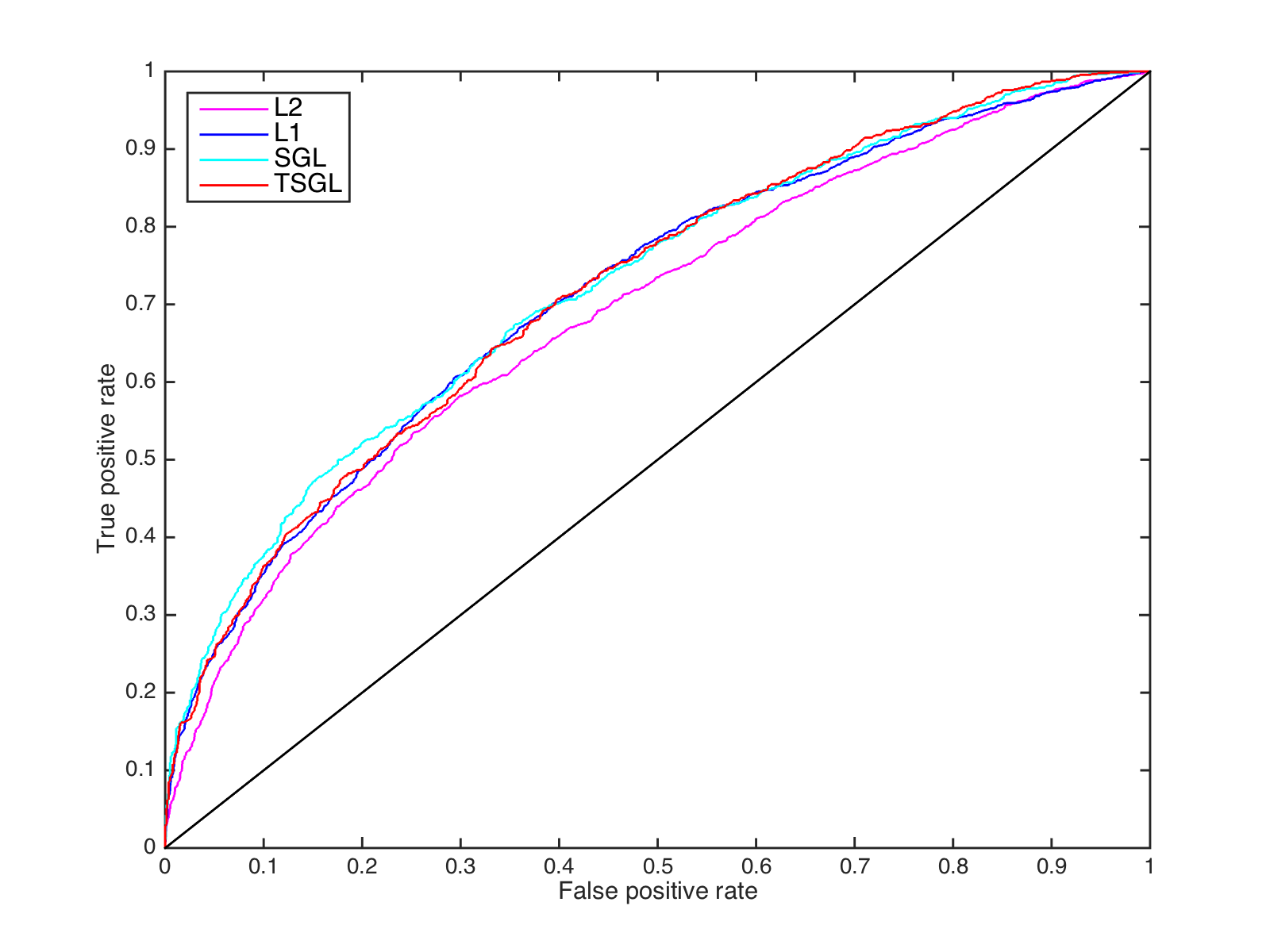}
	\caption{ROC for Different Regularization Strategies}
	\label{fig:roc}
\end{figure}

In the past, researchers have shown that logistic regression typically outperforms other methods for predicting readmissions. Here we compare the performance of different regularization methods discussed in Section~\ref{sec:methodology}. The results are summarized in Table~\ref{tab:regularizationcomparison} and Figure~\ref{fig:roc}.
\begin{table}[htbp]
  \centering
  \begin{tabular}{|c|c|c|c|}
	\hline
	\multirow{2}{*}{Regularization} &
	\multicolumn{2}{c|}{F1 Measure} &
	\multirow{2}{*}{AUC}  \\
	\cline{2-3}
  		& Mean & Std. & \\
	\hline
	l2   &  0.5364 & 0.0044  & 0.6889\\ \hline
	l1   &  0.5343 & 0.0092  & 0.7152\\ \hline
	SGL   &  0.5997 & 0.0027  & 0.7236\\ \hline
	{\bf TSGL}   &  {\bf 0.6487} & {\bf 0.0027}  & {\bf 0.7185}\\ \hline
  \end{tabular}
  \caption{Comparison of Different Regularization Strategies}
  \label{tab:regularizationcomparison}
\end{table}

The results indicate that leveraging the structure in the features (disease codes) results in an improvement in the performance of the logistic regression classifier. However, based on the F1-scores, it is not clear which of the structured regularization methods (SGL or TSGL) is better. 
\subsection{Interpretability of models}
\begin{figure}[htbp]
  \centering
  \begin{subfigure}{0.4\textwidth}
    \centering
    \includegraphics[width=\linewidth]{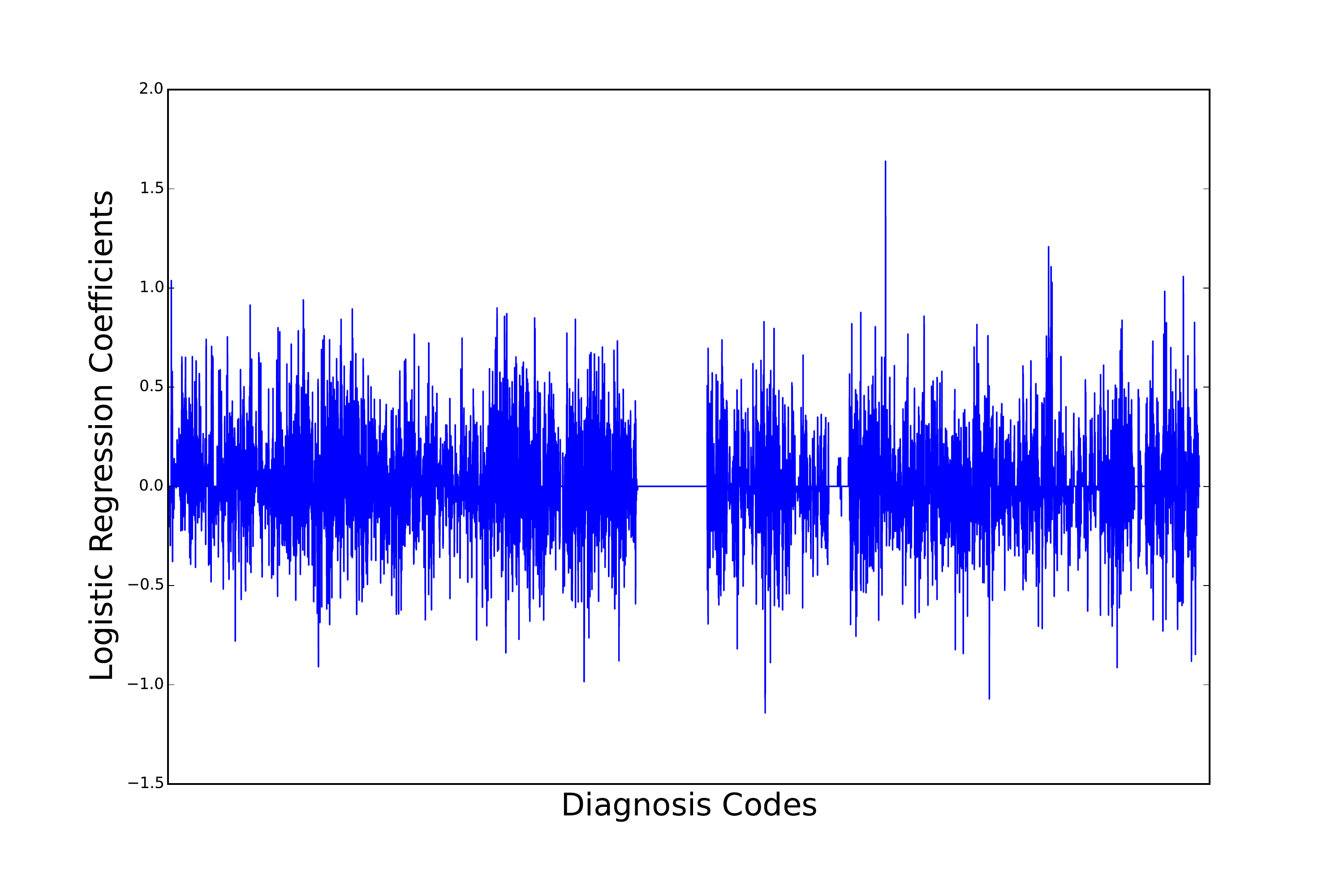}
    \caption{l2}
    \label{fig:l2}
  \end{subfigure}
  \begin{subfigure}{0.4\textwidth}
    \centering
    \includegraphics[width=\linewidth]{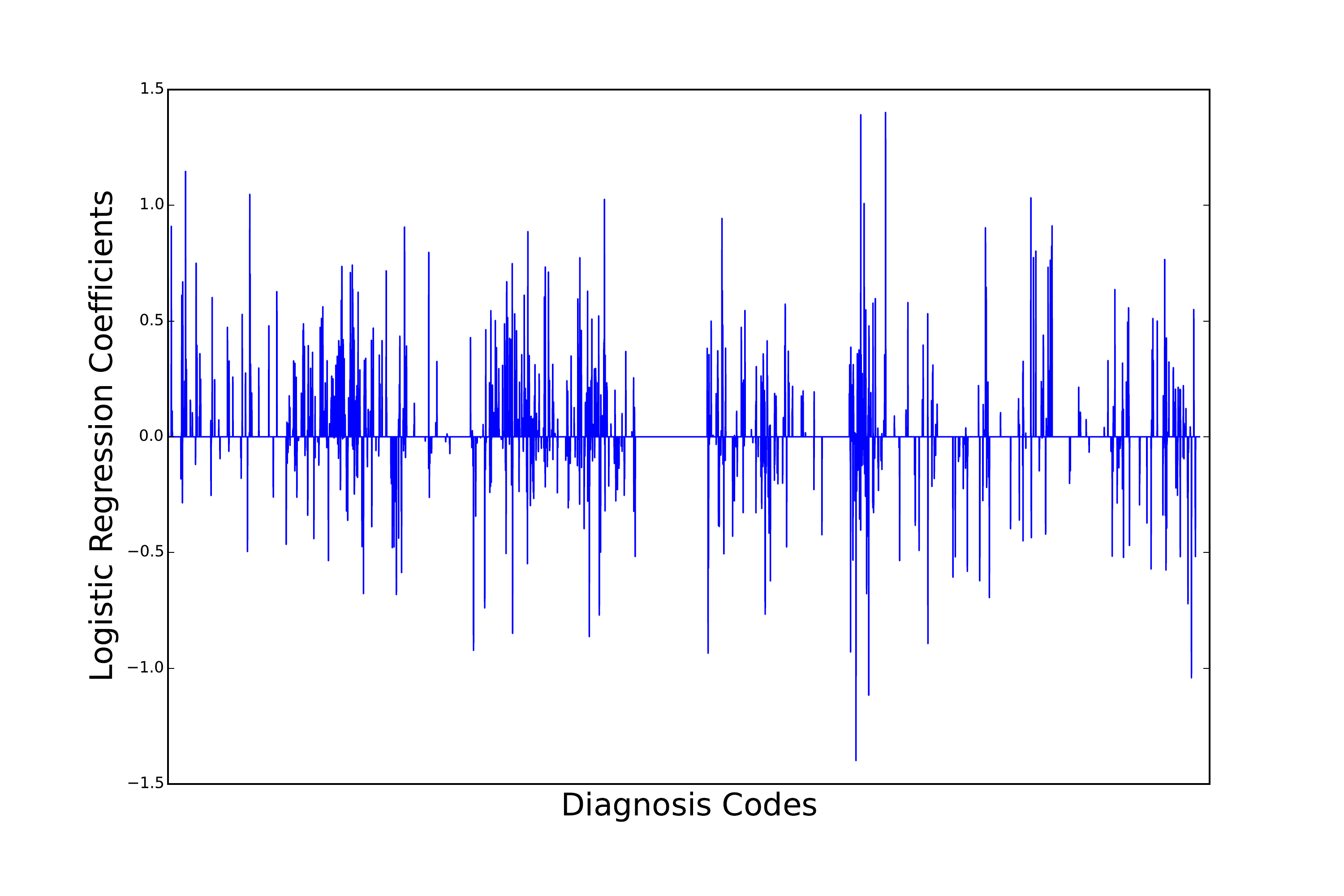}
    \caption{l1}
    \label{fig:l1}
  \end{subfigure}
  \begin{subfigure}{0.4\textwidth}
    \centering
    \includegraphics[width=\linewidth]{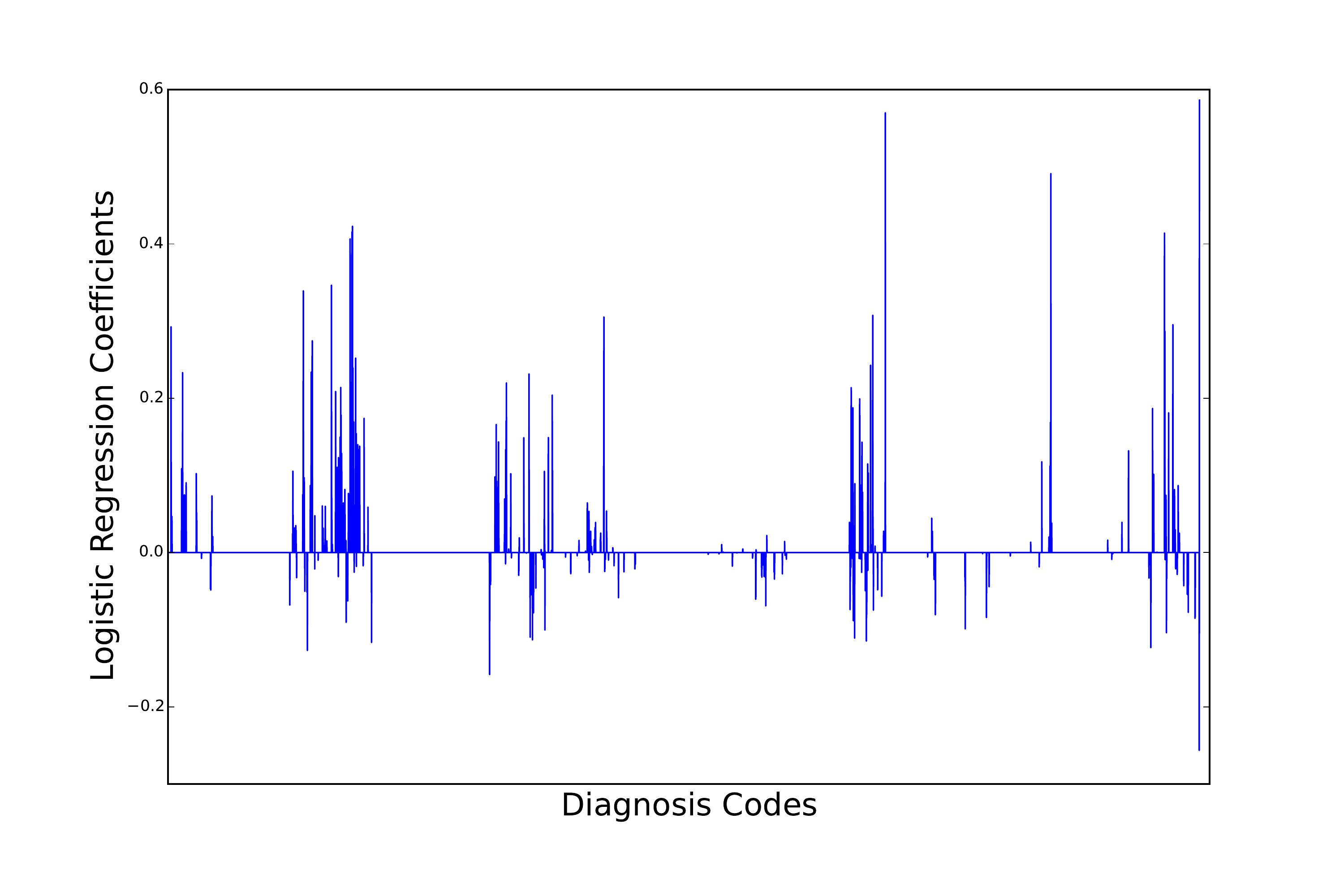}
    \caption{SGL}
    \label{fig:sgl}
  \end{subfigure}
  \begin{subfigure}{0.4\textwidth}
    \centering
    \includegraphics[width=\linewidth]{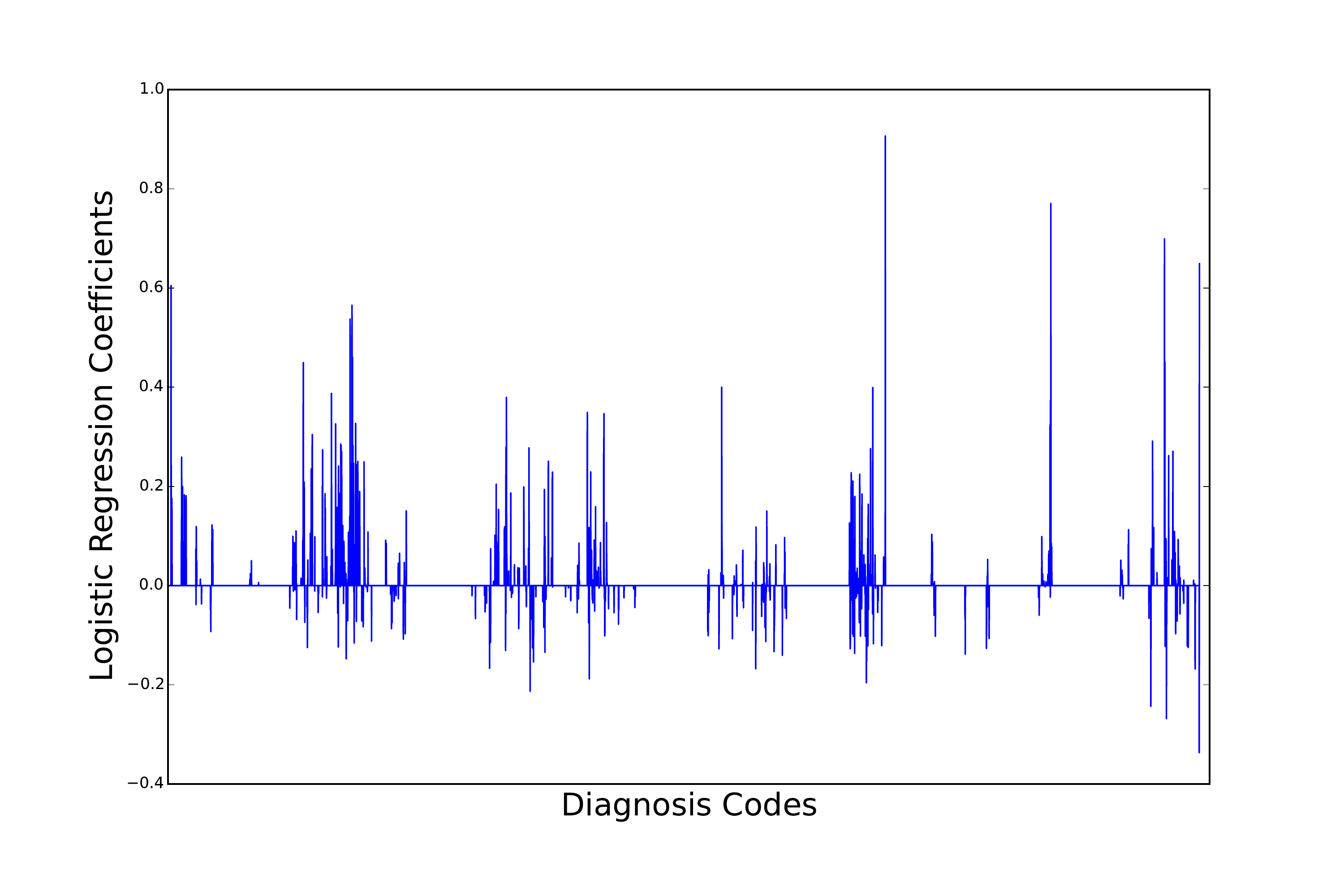}
    \caption{TSGL}
    \label{fig:tsgl}
  \end{subfigure}
  \caption{Distribution of logistic regression coefficient values with different regularizers. The diagnosis codes (features) are arranged in the same order as the ICD-9-CM classification.}
  \label{fig:weights}
\end{figure}

Table~\ref{tab:regularizationcomparison} shows a clear evidence that leveraging the structure in the disease codes allows for a better classifier to predict readmission risk. We now focus on the interpretability of the resulting model under the different regularization mechanisms. The weights for the logistic regression model learnt under the four different types of regularizers is shown in Figure~\ref{fig:weights}. We first note that l2 regularizer, for obvious reasons, does not produce a sparse solution (45\% zero weights), while the other three regularizers induce significant sparsity (l1 - 92\%, SGL - 97\%, and TSGL - 94\%). However, the structured regularizers are able to achieve structured sparsity which is consistent with the ICD-9-CM hierarchy. 

While SGL achieves higher sparsity, the TSGL solution ``aligns'' better with the ICD-9-CM hierarchy. To verify this, we measure the sparsity of the coefficients at different levels of the hierarchy, as shown in Table~\ref{tab:sparsity}. We observe that the TSGL provides better sparsity at higher levels of the hierarchy.
\begin{table}[h]
  \centering
  \begin{tabular}{|l|l|l|}
    \hline
    Level & SGL & TSGL\\
    \hline
    0 & 11537& 11181\\
    1 & 997  & 1021 \\
    2 & 110  & 119\\
    3 & 4    & 6\\
    \hline
  \end{tabular}
    \caption{Number of nodes with all zero coefficients at different levels of the ICD-9-CM hierarchy. Level 0 corresponds to the leafs and level 3 corresponds to the coarsest level.}
    \label{tab:sparsity}
\end{table}

\section{Discussions}
\label{sec:discussions}
Section~\ref{sec:results} shows that leveraging the hierarchical information in the ICD-9-CM classification improves the predictive capability of logistic regression while promoting the structural sparsity to allow for better interpretability. In this section, we study the learnt model from the healthcare perspective. The focus is to understand the factors that impact readmission risk using the coefficients of the model learnt using the TSGL constraint.

As shown in Figure~\ref{fig:tsgl}, the model is well-informed by the ICD-9-CM hierarchy. In all, there are only 437 non-zero coefficients. We choose the top 40 diagnosis codes with highest absolute coefficients. The most important diagnosis codes are listed in Table~\ref{tab:topdiagcodesall}. The codes are grouped by their higher order functions.

The top codes listed in Table~\ref{tab:topdiagcodesall} reveal several valuable insights into the issue of readmission. For instance, while the role of age is understandable (older patients tend to get readmitted more), the impact of the insurance plan supports existing belief that disease management and population health activities provided by MCO would be associated with readmissions~\cite{Hewner:2014}.

The next group of diagnosis codes pertain to infections and complications that happen during the hospital stay of the patients. This information is especially useful for hospitals to identify the key improvement avenues in the hospital operations to reduce the readmission rates. For example, it is well known that post-operative infections result in the patients returning to the hospital shortly after getting discharged. We are able to identify the same issue through this analysis.

While the role of chronic diseases in readmissions is well-understood in the literature, our results indicate that the chronic diseases play less of role, compared to other diagnosis codes. 

However, the most important findings of our study correspond to last three groups of diagnosis codes in Table~\ref{tab:topdiagcodesall}.
Diagnosis codes related to mental health related issues were some of the most important positive factors in predicting readmissions. This is a valuable insight for hospitals in creating post-discharge strategies for such patients. This includes post discharge counseling and home visits.

Another key factor in readmission is substance abuse. In fact, the role of mental health and substance abuse in readmissions has long been studied in the healthcare community~\cite{Lindsey:2007}. Many of the substance abuse issues might not be the direct cause of readmissions but might be indirect indicators of the lack of social and family support to assist patient recovery after the hospital discharge.
\begin{table}[htbp]
\centering
  {\footnotesize
    \begin{tabular}{|p{5in}|}
	\hline
	\rowcolor{mygray} {\bf Age}\\
	\rowcolor{mygray} {\bf Insurance Plan}\\
	\rowcolor{mygray} {\bf Complications of surgery and medical care}\\
	Other unknown and unspecified cause of morbidity and mortality\\
	Hematoma complicating a procedure\\
	Disruption of external operation (surgical) wound\\
	Other postoperative infection\\
	Renal dialysis status\\
	Aortocoronary bypass status\\
	Care involving other specified rehabilitation procedure\\
	Encounter for antineoplastic chemotherapy\\
	Intestinal infection due to Clostridium difficile\\
	\rowcolor{mygray} {\bf Chronic Diseases}\\
	Anemia of other chronic disease\\
	Systolic heart failure, unspecified\\
	Chronic diastolic heart failure\\
	Hypotension, unspecified\\
	Paralytic ileus\\
	Acute kidney failure\\
	Pressure ulcer, lower back\\
	Other ascites\\
	Bacteremia\\
        Hypopotassemia\\
	\rowcolor{mygray} {\bf Mental Disorders}\\
	Paranoid type schizophrenia, chronic with acute exacerbation\\
	Unspecified schizophrenia, chronic\\
	Major depressive affective disorder\\
	Bipolar I disorder\\
	Other personality disorders\\
	Suicidal ideation\\
	\rowcolor{mygray} {\bf Substance Abuse}\\
	Alcohol withdrawal\\
	Acute alcoholic intoxication in alcoholism, unspecified\\
	Other and unspecified alcohol dependence, unspecified\\
	Opioid type dependence, unspecified\\
	Combinations of drug dependence excluding opioid type drug, unspecified\\
	Sedative, hypnotic or anxiolytic abuse, continuous\\
	Long-term (current) use of aspirin\\
	\rowcolor{mygray} {\bf Socio-economic Factors}\\
	Lack of housing\\
	Unspecified protein-calorie malnutrition\\
	\hline
    \end{tabular}
  }
  \caption{Most important diagnosis codes for predicting readmission risk}
  \label{tab:topdiagcodesall}
\end{table}

A direct evidence of the socio-economic effect on readmissions is given by the last set of diagnosis codes which includes lack of housing and malnutrition. These factors had very high positive weights indicating their importance. However, in most existing studies such factors were not considered, primarily because of the lack of relevant data. Through this study we discovery that insurance claims data itself contains elements that can be used to assess the socio-economic background of a patient. 

\section{Analyzing Sub-populations with Chronic Diseases}
\label{sec:subpop}
To further understand the role of various diagnosis codes in predicting readmissions in the context of major chronic diseases, we studied subpopulations of patients suffering from one of 9 diseases listed in Table~\ref{tab:subpopulations}. Most of these diseases have unique symptoms and treatments and many hospitals and other medical facilities are specialized in one of these diseases. Hence, it is important to understand if the important factors for readmission prediction are different from the entire population.

\begin{table}[t]
  \centering
    {\footnotesize
    \begin{tabular}{|l|c|c|}
    \hline
    & \multicolumn{2}{c|}{\bf Number of patients}\\\cline{2-3}
    {\bf Major Disease Group}  &  {\bf Readmission} & {\bf All patients} \\ 
    \hline
    Hyperlipidemia-lipid disorder, LD  &   1448 & 4190 \\ \hline
    Hypertension, HTN  &   2405 & 6407 \\ \hline
    Asthma, ASTH  &    1036 & 2832 \\ \hline
    Chronic obstructive lung Disease, COPD &  1075 & 2798 \\ \hline
    Depression, DEP  &  2268 & 5598 \\ \hline
    Diabetes, DM  &    1267 & 3321 \\ \hline
    Coronary Artery Disease, CAD  &   300 & 658 \\ \hline
    Heart Failure, HF  &   736 & 1692 \\ \hline
    Chronic Kidney Disease, CKD   &  528 & 1327 \\ \hline  
  \end{tabular}}
  \caption{Details of subpopulations suffering from a major chronic disease}
  \label{tab:subpopulations}
  \end{table}
\begin{table}[h]
    \centering
    {\footnotesize
  \begin{tabular}{|l|c|c|}
    \hline
    Major disease group  & F1 Measure &  Std.\\
    \hline
    LD  &  0.5631 & 0.0091 \\ \hline
    HTN  &  0.5966 & 0.0057 \\ \hline
    ASTH  &  0.5815 & 0.0110 \\ \hline
    COPD  &  0.6031 & 0.0183 \\ \hline
    DEP  &  0.6098 & 0.0050 \\ \hline
    DM &  0.6200 & 0.0183 \\ \hline
    CAD  &  0.6588 & 0.0288 \\ \hline
    HF &  0.6457 & 0.0250 \\ \hline
    CKD  &  0.6239 & 0.0209 \\ \hline  
  \end{tabular}
}
  \caption{Performance of TSGL based logistic regression classifier on chronic disease subpopulations}
  \label{tab:subpopulationresults}
 \end{table}
Table~\ref{tab:subpopulationresults} shows the average performance of the logistic regression classifier (using cross-validation) on subpopulations corresponding each disease code. The subpopulations are created by considering only those patients who have had at least one admission for a given chronic disease. The results match the values obtained for the entire population.
Next, we study the diagnosis codes which are most important in predicting readmission risk. Generally, we observe that while a few codes are common to all disease types, there are certain disease codes which are exclusively unique to each of the subtype. The diagnosis codes that have a strong impact on readmissions, independent of the type of chronic disease the patient suffers from, include postoperative infection and suicidal ideation. 
\begin{table}[htbp]
\centering
  {\footnotesize
    \begin{tabular}{|p{5in}|}
    	
	\hline
	\rowcolor{mygray} {\center \bf Entire population}\\
	Age\\
         Insurance Plan\\
         Suicidal Ideation\\
         Other Postoperative Infection\\
	 \hline
	\rowcolor{mygray} {\center \bf Diabetes}\\
      		Unspecified transient cerebral ischemia\\
      		Abnormality of gait \\
      	Schizophrenic disorders, residual type, chronic\\
      	Methicillin susceptible Staphylococcus aureus septicemia\\
      			Hematemesis\\
      	Chronic hepatitis C without mention of hepatic coma\\
\hline	
	\rowcolor{mygray} {\center \bf Hyperlipidemia}\\
	Persistent vomiting\\
      	Diverticulitis of colon (without mention of hemorrhage)\\
    \hline	
	\rowcolor{mygray} {\center \bf Hypertension}\\
	Candidiasis of mouth\\
				Anemia of other chronic disease \\
			  	Injury of face and neck\\
\hline	
	\rowcolor{mygray} {\center \bf Coronary Artery Disease}\\
	Chronic obstructive asthma, unspecified\\
      	Asthma,unspecified type, unspecified\\
      	Diabetes with peripheral circulatory disorders\\
      		Urinary tract infection, site not specified\\
      		Epistaxis\\
      		Chronic kidney disease, Stage III (moderate)\\
      	Cerebral embolism with cerebral infarction\\
	\hline	
	\rowcolor{mygray} {\center \bf Chronic Kidney Disease}\\
	Hydronephrosis\\
				Diabetes mellitus\\
					Orthostatic hypotension\\
				Other diseases of lung\\
				Ventricular fibrillation\\
				Acute respiratory failure\\
	\hline	
	\rowcolor{mygray} {\center \bf Chronic Obstructive Lung Disease}\\
	Secondary malignant neoplasm of bone and bone marrow\\
	Thrombocytopenia, unspecified\\
	Unspecified acquired hypothyroidism\\
	Acute pancreatitis\\
	\hline	
      \end{tabular}}
      \caption{Most important diagnosis codes unique for few selected chronic disease specific subpopulations. Codes that are related to the same disease are ignored.}
      \label{tab:subprobcodes}
    \end{table}
    On the other hand, for every chronic disease, there are certain diagnosis codes that are unique, i.e., they are not a factor in any other chronic disease. Such codes can be significant for organizations that specialize in such diseases. Some of these codes are indicative of other chronic diseases or comorbidities. For example, for Diabetes patients, the presence of {\em transient cerebral ischemia} is an important predictor of readmission risk. Similarly, the presence of {\em pancreatic disease} codes in a COPD patient is an important risk factor. Interestingly, the unique diagnosis codes identified for each of the major chronic diseases primarily contain other chronic disease related codes, thereby indicating a strong impact of comorbidity in readmissions. 

\section{Conclusions}
\label{sec:conclusions}
In the last decade, there have been numerous studies that link factors pertaining to a patient's hospital stay to the risk of readmission. However most studies have been on a focused cohort, limited to one or few hospitals. However, we show here that similar results can be achieved using claims data, which has fewer elements but provides a large population coverage; the entire state of New York for this study. Even with the large volume of data, the predictive algorithms are not accurate enough ($\sim 0.60$ F1-score) to be used as decision making tools. However, model interpretation can reveal insights which can inform the strategies for reducing and/or eliminating readmissions. 

A patient's disease history is typically expressed using diagnosis codes, which can take as many as 18000 possible values, with many more possibilities in the next generation ICD-10 disease classification. With so many possible features, ensuring model interpretability is a challenge. However, using structured sparsity inducing models, such as the tree sparse group LASSO, used in this paper, one can ensure that the truly important factors can be identified. In this case study, we discover several such interesting factors.

In particular, we conclude that while in-hospital events such as infections are important, behavioral factors such as mental disorders and substance abuse and socio-economic factors, such as lack of housing or malnutrition at home are equally important. Targeted strategies, such as phone calls and home visits, will need to be developed to handle such situations. In Section~\ref{sec:subpop}, we analyze subpopulations specific to chronic diseases and show that similar methodology can reveral disease specific factors for readmissions.

{\small
\section*{Acknowledgements}
This material is based in part upon work supported by
the National Science Foundation under award number CNS
- 1409551. 
}

%
%

\bibliographystyle{splncs04}
\bibliography{refs}

\end{document}